\pdfoutput=1

\documentclass[11pt]{article}

\usepackage[preprint]{acl}
\usepackage{markdown}
\usepackage{times}
\usepackage{latexsym}

\usepackage[T1]{fontenc}

\usepackage[utf8]{inputenc}

\usepackage{microtype}

\usepackage{inconsolata}

\usepackage{graphicx}

\usepackage{hyperref}       
\usepackage{url}            
\usepackage{booktabs}       
\usepackage{amsfonts}       
\usepackage{nicefrac}       
\usepackage{microtype}      
\usepackage{lipsum}
\usepackage{fancyhdr}       
\usepackage{graphicx}       
\usepackage{float}
\usepackage{multirow}
\usepackage[utf8]{inputenc}
\graphicspath{{media/}}     
\usepackage{siunitx} 
\usepackage{gensymb} 
\usepackage[most]{tcolorbox}

\usepackage{markdown}
\usepackage{tabularx}
\usepackage{xcolor}
\usepackage{hyperref}
\usepackage{multirow}
\usepackage{listings}
\usepackage{longtable}
\usepackage{bbding}
\usepackage{enumitem}
\usepackage{amssymb}
\usepackage{wrapfig}
\usepackage{array}
\hypersetup{
    colorlinks=true, 
    linkcolor=blue, 
    filecolor=magenta, 
    urlcolor=cyan, 
    citecolor=green 
}

\newtcolorbox{prompt}[1]{
    enhanced,
    colback=gray!20,
    colframe=black,
    boxrule=0.3pt,
    arc=3mm,
    left=2pt,
    right=2pt,
    boxsep=3pt,
    fonttitle=\small\bfseries,
    title=#1,
    fontupper=\scriptsize
}

\definecolor{lightgray}{gray}{0.9}
\DeclareUnicodeCharacter{FF08}{\textlparen} 
\DeclareUnicodeCharacter{FF09}{\textrparen} 

\title{FlightGPT: Towards Generalizable and Interpretable UAV Vision-and-Language Navigation with Vision-Language Models}

\author{%
\parbox{\linewidth}{\centering%
\textbf{Author Name$^{1}$\thanks{Equal Contribution}} and \textbf{Author Name$^{1,2}$} \\[2ex] 
$^1$Institution Name \quad $^2$Institution Name \\[2ex] 
$^3$Institution Name \quad $^4$Institution Name \\[2ex] 
$^5$Institution Name \quad $^6$Institution Name \\[2ex] 
$^7$Institution Name \\[2ex] 
}%
}

\author{%
\parbox{\linewidth}{\centering%
\textbf{Hengxing Cai$^{1,2}$\thanks{Equal Contribution}}, \textbf{Jinhan Dong$^{2,3}$\footnotemark[1]}, \textbf{Jingjun Tan$^{1}$}, \textbf{Jingcheng Deng$^4$}, \textbf{Sihang Li$^{2}$}, \\[0.5ex] \textbf{Zhifeng Gao$^2$},  \textbf{Haidong Wang$^1$},  \textbf{Zicheng Su$^5$}, \textbf{Agachai Sumalee$^6$} and \textbf{Renxin Zhong$^{1}$\thanks{Corresponding author}} \\[2ex]
$^1$School of Intelligent Systems Engineering, Sun Yat-Sen University \quad $^2$DP Technology \quad $^3$Beijing University Of Posts and Telecommunications \\ $^4$Institute of Computing Technology, Chinese Academy of Sciences \quad $^5$Tongji University \\ $^6$School of Integrated Innovation, Chulalongkorn University \\[5ex]
}%
}

\begin{document}
\maketitle
\begin{abstract}
Unmanned Aerial Vehicle (UAV) Vision-and-Language Navigation (VLN) is vital for applications such as disaster response, logistics delivery, and urban inspection. 
However, existing methods often struggle with insufficient multimodal fusion, weak generalization, and poor interpretability.
To address these challenges, we propose \textbf{FlightGPT}, a novel UAV VLN framework built upon Vision-Language Models (VLMs) with powerful multimodal perception capabilities.
We design a two-stage training pipeline: first, Supervised Fine-Tuning (SFT) using high-quality demonstrations to improve initialization and structured reasoning; then, Group Relative Policy Optimization (GRPO) algorithm, guided by a composite reward that considers goal accuracy, reasoning quality, and format compliance, to enhance generalization and adaptability.
Furthermore, FlightGPT introduces a Chain-of-Thought (CoT)-based reasoning mechanism to improve decision interpretability.
Extensive experiments on the city-scale dataset CityNav demonstrate that FlightGPT achieves state-of-the-art performance across all scenarios, with a 9.22\% higher success rate than the strongest baseline in unseen environments.
Our implementation is publicly available\footnote{\url{https://github.com/Pendulumclock/FlightGPT}}.
\end{abstract}


\section{Introduction}
With the rapid advancement of Unmanned Aerial Vehicles (UAV) technology, vision-and-language navigation (VLN) has emerged as a critical capability for UAV applications~\cite{fan2022aerial, li2025navblip, sautenkov2025uav, wu2024vision}. 
Specifically, UAV VLN involves the ability to comprehend and integrate natural language instructions with visual observations, enabling UAVs to plan and execute flight missions in complex and dynamic real-world environments~\cite{wang2024towards}.
This capability has demonstrated significant value across a variety of practical scenarios~\cite{wang2024multimodal}. 
For example, during disaster relief operations, UAVs can rapidly identify disaster-affected areas and plan safe routes based on rescue instructions, thereby improving the effectiveness of search and rescue missions~\cite{estrada2019uses}. 

Despite numerous methods being developed for UAV VLN task --- such as sequence-to-sequence (Seq2Seq)~\cite{2018Speaker}, Cross-Modal Attention (CMA)~\cite{liu2023aerialvlnvisionandlanguagenavigationuavs}, and Map-based Goal Predictors (MGP)~\cite{lee2024citynavlanguagegoalaerialnavigation} --- several critical challenges remain in practical applications.

\textbf{Insufficient multimodal information fusion.} Existing methods often perform simple concatenation or shallow fusion of image and text inputs, lacking effective integration of deep semantic understanding and visual perception. 
Therefore, navigation strategies are prone to misinterpretation of complex instructions and perception errors, leading to suboptimal navigation performance.

\textbf{Weak generalization and poor dynamic adaptability.} Existing models typically rely heavily on the training environment and lack generalization capabilities in Out-of-Distribution (OOD) environments. 
When encountering unfamiliar environments or dynamic obstacles, their navigation performance degrades significantly, making reliable execution challenging.

\textbf{Limited Interpretability of Navigation Decisions.} Most current approaches directly output navigation decisions without providing clear intermediate reasoning steps. The decision-making logic is opaque to users, making it difficult to diagnose errors or refine navigation strategies, which limits the system's reliability and maintainability.

\begin{figure*}[t]
  \centering
  \includegraphics[width=0.97\textwidth]{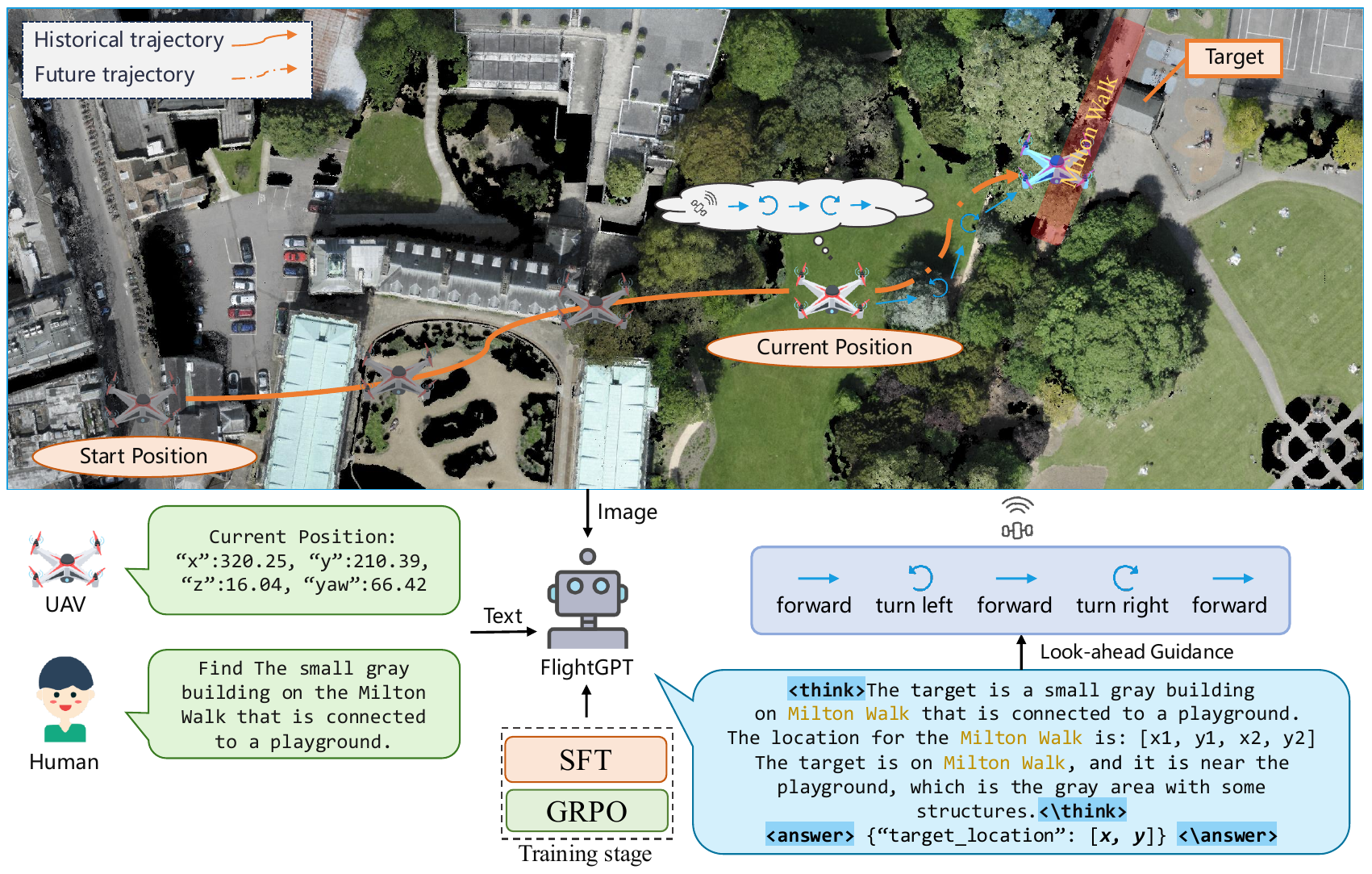} 
  \vspace{-8pt}
  \caption{Workflow of FlightGPT for UAV VLN. FlightGPT takes multimodal input comprising a semantic map image and a natural language instruction, performs Chain-of-Thought reasoning to infer the target location, which is used for subsequent executable actions.}
  \label{fig:inference}
\end{figure*}

To address these challenges, we propose \textbf{FlightGPT}, a novel UAV VLN framework, as illustrated in Fig.~\ref{fig:inference}.
The system is built upon Vision-Language Models (VLMs) to support multimodal understanding, strong generalization and adaptability, and interpretable reasoning.
The design of FlightGPT focuses on three techniques:

\textbf{VLM-based multimodal integration.} Utilizing the capacity of VLMs, visual and textual inputs are effectively integrated to enhance multimodal perception and understanding.

\textbf{Two-stage training pipeline.}
A two-stage training pipeline is designed, starting with supervised fine-tuning (SFT) on high-quality demonstrations to warm up the policy, followed by reinforcement learning (RL) with a task-specific reward designed for UAV VLN to improve model generalization.

\textbf{Chain-of-Thought based reasoning module.} A structured reasoning mechanism is introduced using explicit \texttt{<think>} / \texttt{<answer>} tags, forming a Chain-of-Thought (CoT) reasoning process. 
This design enables the model to “think before acting” and improves reasoning quality.

The main contributions of this work are summarized as follows:
\begin{enumerate}
    \item We leverage an end-to-end VLM that effectively integrates visual and textual inputs for enhanced multimodal comprehension.

    \item A two-stage training pipeline is developed, where SFT helps accelerate convergence and stabilize training, followed by RL to enhance the model’s generalization and adaptability.

    \item A CoT reasoning mechanism is introduced to improve the model’s reasoning quality, resulting in reasoning processes that are more complete, coherent, and fluent.

    \item We evaluate FlightGPT on CityNav, a large-scale benchmark based on real-world urban environments. The model achieves state-of-the-art performance, and demonstrates strong generalization.
\end{enumerate}

\section{Related Work and Motivation}

\subsection{Evolution of UAV Vision-and-Language Navigation}

UAV VLN plays a key role in enabling intelligent flight in complex environments, and its research has undergone continuous evolution.
Early UAV VLN approaches adopted Seq2Seq models that encoded language instructions into fixed representations for action generation~\cite{2018Speaker}.
CMA mechanisms were later proposed to enhance alignment between navigation instructions and visual observations~\cite{liu2023aerialvlnvisionandlanguagenavigationuavs}, while the Self-Monitoring model incorporated auxiliary progress estimation to support self-correction during navigation~\cite{2019Self}.
With the rise of Transformer architectures, pretrained models such as VLN-BERT~\cite{2021VLNBERT} were introduced, employing a multimodal BERT structure to integrate language and visual trajectories.
Alongside method development, UAV VLN benchmarks have also evolved.
AerialVLN~\cite{liu2023aerialvlnvisionandlanguagenavigationuavs} introduced a high-fidelity 3D simulation environment for language-guided flight, while CityNav~\cite{lee2024citynavlanguagegoalaerialnavigation} provides a city-scale dataset with GPS, imagery, and natural language, increasing task diversity and evaluation complexity.
These developments have promoted the intelligent evolution of UAV VLN technologies and the standardization of benchmark datasets.

\subsection{Vision-Language Models for Multimodal Perception in Navigation}
VLMs, pretrained on large-scale image-text corpora, have demonstrated strong capabilities in unifying visual and linguistic modalities, making them increasingly relevant to navigation tasks that demand rich semantic perception.
Early models such as UNITER~\cite{chen2020uniteruniversalimagetextrepresentation} aligned image and text features in a joint embedding space, while CLIP~\cite{radford2021learning} introduced contrastive learning for open-vocabulary visual recognition, greatly improving the generalization of multimodal representations.
Recent VLMs like GPT-4V~\cite{openai2024gpt4technicalreport}, Gemini 1.5~\cite{geminiteam2024gemini15unlockingmultimodal}, and Qwen2-VL~\cite{wang2024qwen2vlenhancingvisionlanguagemodels} further expand this capability, enabling unified interfaces for vision-language reasoning and decision support.
In navigation contexts, researchers have preliminarily shown that VLMs can directly process multimodal inputs to generate navigation trajectories or structured subtasks~\cite{wang2024towards}.
This ability to unify visual perception with language understanding positions VLMs as a promising foundation for bridging high-level task interpretation and low-level action control in navigation.

\subsection{Reinforcement Learning for Enhancing Generalization in Navigation}
RL has emerged as an effective mechanism for enhancing both the reasoning capabilities and generalization performance of large language models (LLMs) and embodied agents.
DeepSeek-R1~\cite{deepseekai2025deepseekr1incentivizingreasoningcapability} applies large-scale RL to optimize chain-of-thought reasoning in language models, yielding strong performance in complex tasks such as mathematical problem solving and code generation.
Beyond static reasoning, RL has also been leveraged to improve model adaptability in interactive settings.
GROOT~\cite{zhu2023learninggeneralizablemanipulationpolicies} trains general-purpose agents in 3D environments through end-to-end RL, demonstrating the ability to generalize across diverse manipulation tasks via object-centric representations.
These studies highlight the dual role of RL: not only reinforcing structured reasoning in LLMs, but also enhancing their robustness and transferability across dynamic and multi-task environments.
Such capabilities are particularly valuable for UAV VLN, where agents must interpret diverse language inputs and adapt to complex, ever-changing visual contexts.


\subsection{Limitations of Existing Work and Motivation for FlightGPT}
While recent advances in VLMs have improved multimodal perception and RL has shown strong potential in enhancing policy generalization, their application in UAV VLN remains limited due to challenges in action reliability and training stability.
To address these limitations, we propose FlightGPT, a unified framework that combines the perceptual capabilities of VLMs with the adaptive learning strengths of RL to provide a more generalizable and effective solution for UAV VLN.

\section{Method}

\subsection{Problem Formulation} 
We focus on the task of UAV VLN, which requires the UAV to reach a designated target in a three-dimensional environment. The navigation process is guided by both a natural language description of the target and the UAV’s visual perception of its surroundings.
Specifically, each task can be formalized as a triplet $(I, D, E)$, where:
\begin{itemize}[leftmargin=*]
    \item $I$ denotes the initial state of the agent, including its position and heading angle;
    \item $D$ represents a natural language description of the target, typically including details about the target and its surrounding landmarks;
    \item $E$ refers to a three-dimensional environment with realistic spatial layouts and rich geographic semantics, where the UAV can access various perceptual inputs such as key landmarks, as well as RGB/depth maps from its first-person viewpoint.
\end{itemize}



The agent accomplishes the navigation task by executing a sequence of discrete actions, including \textbf{forwad}, \textbf{turning left}, \textbf{turning right}, \textbf{ascend}, \textbf{descend}, \textbf{stop}.
When the agent determines that it has arrived near the target, it can choose the \textbf{stop} action. 
The navigation is considered successful if the final position of the UAV is within a predefined distance threshold (e.g., 20 meters) from the target.

\subsection{FlightGPT}

\subsubsection{System Overview}


\begin{figure*}[t]
  \centering
  \includegraphics[width=0.97\textwidth]{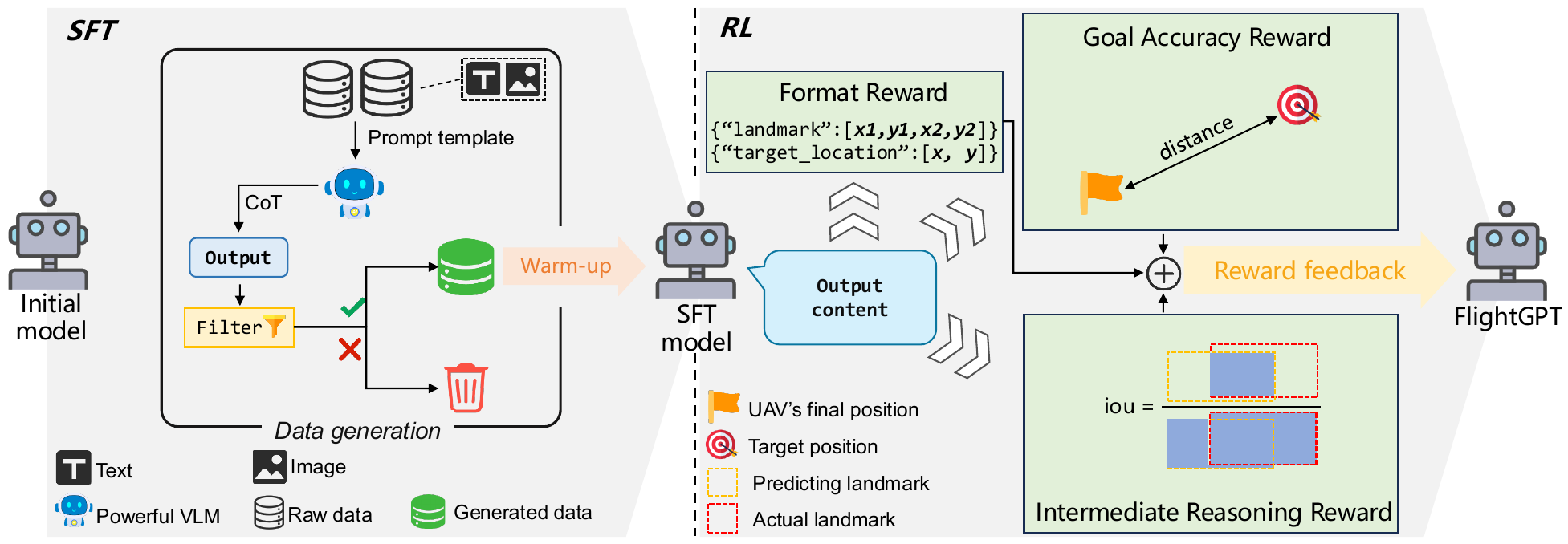} 
  \vspace{-8pt}
  \caption{The two-stage training pipeline of FlightGPT. The pipeline consists of a supervised fine-tuning (SFT) stage using CoT-annotated data generated by a powerful VLM, followed by reinforcement learning (RL) with composite rewards, including goal accuracy, intermediate reasoning, and format compliance.}
  \label{fig:training}
\end{figure*}

Fig.~\ref{fig:inference} illustrates the inference process of FlightGPT when executing the UAV VLN task, which consists of the following steps:

\textbf{1. Input Acquisition.} The system collects inputs from the environment, including a semantic map (annotated with the UAV's current position, heading angle, first-person field of view, and known landmark information) and a textual description (containing the UAV's current position and a natural language description of the target).

\textbf{2. Reasoning and Target Prediction.} Following the paradigm adopted by several existing methods (e.g., Seq2Seq, CMA, MGP), we adopt a sequential workflow that first predicts the target location and then plans the navigation actions. FlightGPT generates a structured reasoning process and outputs a prediction of the target location.

\textbf{3. Action Planning.} Following the approach proposed in AerialVLN~\cite{liu2023aerialvlnvisionandlanguagenavigationuavs}, we incorporate a look-ahead mechanism into our system, enabling the simulation of future trajectories for generating executable actions.

\textbf{4. Environment Interaction.} The UAV executes the planned actions in the environment and updates its state.

This process is iteratively repeated until the UAV either executes a \textbf{stop} action or reaches a predefined maximum number of iterations.

Inspired by the training process of DeepSeek-R1~\cite{deepseekai2025deepseekr1incentivizingreasoningcapability}, we design a two-stage training pipeline to equip \textit{FlightGPT} with the aforementioned capabilities, as illustrated in Fig.~\ref{fig:training}.
\begin{itemize}[leftmargin=*]
    \item \textbf{Stage 1: SFT.}  
    We use a strong model to generate training data that includes CoT-style reasoning processes. After selecting high-quality samples, we perform SFT to train the initial model. 
    This stage is designed to provide the model with a solid initialization and to endow it with the ability to perform structured reasoning.
    
    \item \textbf{Stage 2: RL.}  
    Building upon the SFT stage, we introduce a reinforcement learning phase based on the Group Relative Policy Optimization (GRPO) algorithm, guided by carefully designed rewards. Specifically, we define three types of rewards: Goal Accuracy Reward, Intermediate Reasoning Reward, and Format Reward. This stage aims to enhance the model's generalization ability and robustness in complex and dynamic environments.
\end{itemize}

\subsubsection{SFT for Warm-up} 



While VLMs exhibit strong multimodal understanding, they still struggle with perception and decision-making tasks in complex and dynamic environments, such as UAV VLN. Meanwhile, RL often suffers from unstable convergence when trained from scratch.
To address these challenges, we introduce a SFT stage that leverages high-quality demonstrations to warm up the model, providing a solid initialization for subsequent RL optimization.

\paragraph{Input, Prompt, and Output Design} 
\label{sec:input-output-prompt}
To enable structured output generation and strong reasoning capabilities, we design the input, prompt, and output format during the SFT stage as follows:

\textbf{Input}
The input is composed of two parts: (1) Semantic Map, which is a map annotated with the UAV's current position and heading angle, the first-person view region projected on the map, and the locations of known landmarks; and (2) Textual Information, which describes the UAV's current state information, including its position and heading angle, along with a natural language description of the target.


\textbf{Prompt}
To enable structured reasoning and enhance interpretability, we design a prompt template that explicitly induces a CoT style reasoning process. The prompt includes a detailed system message outlining the UAV’s role and mission objective, along with structured descriptions of both the semantic map and the textual target instruction. It guides the model to reason step-by-step within dedicated <think> tags—covering semantic understanding of the target, landmark identification, and spatial inference, and to produce the final location prediction within <answer> tags.
This CoT-style prompting not only improves reasoning completeness but also provides an interpretable output format that facilitates model debugging and performance analysis.
The full prompt template is provided in Appendix~\ref{app:prompt}.

\textbf{Output}
The output consists of two components:
\begin{itemize}[leftmargin=*]
    \item \texttt{<think> ... </think>}: The model's intermediate reasoning process, which may include understanding the target, recognizing landmarks, and inferring spatial relationships.
    \item \texttt{<answer> ... </answer>}: The final predicted target location, which are used for generating subsequent executable actions.
\end{itemize}

\paragraph{Data Generation} 
Due to the lack of reasoning datasets tailored for UAV VLN tasks, we adopt the Qwen2.5-VL-32B model to automatically generate the training data required for the SFT stage. 
Without any additional fine-tuning, we compared several open-source and closed-source VLMs, and Qwen2.5-VL-32B demonstrated the best performance under the same settings. Therefore, we select it as our data generator. 
Specifically, we provide Qwen2.5-VL-32B with the input and prompt template described in Section~\ref{sec:input-output-prompt}, guiding it to output both the reasoning process and the final prediction. 
To ensure the quality of the training data, we introduce the following filtering and augmentation mechanisms: (1) discard samples with abnormal output formats; (2) discard samples where the predicted location is more than 20 meters away from the ground truth; and (3) for retained samples, replace the target location predicted by Qwen2.5-VL-32B with the ground truth.


\paragraph{Training Strategy} 

The training objective is next-token prediction, where the model predicts the next token based on the given input and previously generated context, proceeding token-by-token until the entire output sequence is completed.

\subsubsection{RL for Generalization} 
Although the model acquires preliminary abilities in visual-language understanding and reasoning through the SFT stage, it still lacks the adaptability required for complex and dynamic environments—particularly in terms of generalization to unseen scenarios.
To address this, we adopt the GRPO algorithm to perform policy optimization using pre-collected simulated data, based on the multimodal input and prompt templates defined in Section~\ref{sec:input-output-prompt}.

To jointly improve final navigation accuracy, reasoning quality, and output format consistency, we design a composite reward system consisting of the following three components:

\paragraph{Goal Accuracy Reward.}
The accuracy of the predicted target location is a key indicator of the system's effectiveness. Let the UAV's predicted position be $\mathbf{\hat{p}} = (\hat{p}_x, \hat{p}_y)$ and the ground truth be $\mathbf{p^*} = (p^*_x, p^*_y)$. We define the reward based on their distance:
\begin{center}
\resizebox{\linewidth}{!}{$
R_{\text{goal}} =
\begin{cases}
1, & \text{if } d(\mathbf{\hat{p}}, \mathbf{p^*}) \leq d_{\text{success}} \\
\exp\left( -\dfrac{d(\mathbf{\hat{p}}, \mathbf{p^*}) - d_{\text{success}}}{\tau} \right), & \text{if } d_{\text{success}} < d(\mathbf{\hat{p}}, \mathbf{p^*}) \leq d_{\text{cutoff}} \\
0, & \text{otherwise}
\end{cases}
$}
\end{center}
where:
\begin{itemize}[leftmargin=*]
    \item The Euclidean distance $d(\mathbf{\hat{p}}, \mathbf{p^*})$ is defined as:
    \[
    d(\mathbf{\hat{p}}, \mathbf{p^*}) = \sqrt{(\hat{p}_x - p^*_x)^2 + (\hat{p}_y - p^*_y)^2}.
    \]
    \item $d_{\text{success}} = 20$ meters: threshold for task success;
    \item $d_{\text{cutoff}} = 80$ meters: upper limit beyond which no reward is given;
    \item $\tau = 100$: decay temperature controlling the sharpness of the exponential drop-off.
\end{itemize}

This reward encourages the model to generate target location that are closer to the ground truth, thereby improving success rate.

\paragraph{Intermediate Reasoning Reward.}
Providing guidance for intermediate reasoning steps is critical for enhancing multi-step navigation performance. 
In our task, we leverage landmarks as key intermediate signals to encourage effective reasoning during the \texttt{<think>} stage.
Specifically, we introduce a reward based on the Intersection over Union (IoU) between the predicted landmark bounding box $\hat{B}$ and the ground-truth bounding box $B$. 
The reward is defined as:\[R_{\text{IoU}} = \frac{\text{Area}(B \cap \hat{B})}{\text{Area}(B \cup \hat{B})}\]
If the model fails to output a valid bounding box, we set $R_{\text{IoU}} = 0$.
This mechanism incentivizes spatial reasoning before location prediction and contributes to more stable and interpretable intermediate representations.


\paragraph{Format Reward.}
To ensure the model generates structured outputs, we introduce a format compliance reward. 
This reward encourages the model to produce outputs that follow a predefined format, with both the reasoning and action sections clearly presented and containing the required information.

\begin{itemize}[leftmargin=*]
    \item If the output includes both \texttt{<think>} and \texttt{<answer>} tags properly enclosing the reasoning and answer segments, a reward of $+0.5$ is given;
    \item If a \texttt{"landmark\_bbox"} field in the format \texttt{[x1, y1, x2, y2]} is successfully extracted within the \texttt{<think>} tag, an additional $+0.25$ is granted;
    \item If a \texttt{"target\_location"} field in the format \texttt{[x, y]} is successfully extracted within the \texttt{<answer>} tag, another $+0.25$ is granted.
\end{itemize}

This reward helps stabilize the model's output structure, facilitating controllability and enabling downstream execution or interpretation.

\paragraph{Overall Reward.}
The total reward used for policy optimization is the sum of the three components described above:
\[
R_{\text{total}} = R_{\text{goal}} + R_{\text{IoU}} + R_{\text{format}}
\]
\section{Experiments}
\subsection{Experimental Settings}
\subsubsection{Dataset}

In this study, we utilize the CityNav~\cite{lee2024citynavlanguagegoalaerialnavigation} dataset, a high-quality benchmark specifically designed for city-scale UAV VLN tasks. 
CityNav comprises 32,637 human demonstration trajectories across 5,850 target objects, constructed on top of 3D urban scans from the SensatUrban dataset. 
It covers two real-world cities, Birmingham and Cambridge, with a total area of approximately 4.65 km², providing rich geographic semantics and diverse navigation scenarios. 
The dataset is publicly available under the MIT License, enabling free use for research purposes.




\subsubsection{Evaluation Metrics}
Following the standard evaluation protocol established by CityNav, four metrics are used to evaluate performance:
\begin{itemize}[leftmargin=*]
    \item \textbf{Navigation Error (NE)}: The Euclidean distance between the agent's final position and the ground-truth location. Lower NE indicates better localization accuracy.

    \item \textbf{Success Rate (SR)}: The percentage of episodes in which the agent stops within 20 meters of the target location.

    \item \textbf{Oracle Success Rate (OSR)}: The proportion of episodes where the agent, at any point during navigation, gets within 20 meters of the target, regardless of whether it stops.

    \item \textbf{Success weighted by Path Length (SPL)}: A metric that adjusts SR by penalizing unnecessarily long paths, encouraging efficient navigation.
\end{itemize}

These metrics jointly reflect the agent's goal-reaching accuracy, path efficiency, and overall navigation robustness.

\subsubsection{Baseline Models}
We conduct evaluations of FlightGPT against a diverse set of representative baselines, including Random, Seq2Seq, CMA, MGP, GPT-4o, Qwen2.5-VL (7B / 32B), and LLaMA-3.2-11B-Vision. Brief introductions for all baselines are provided in Appendix~\ref{app:baselines}.


\subsubsection{Model and Training Configuration}
We adopt Qwen2.5-VL-7B as the base model and optimize it using a two-stage pipeline. For SFT, 1,872 samples were collected and filtered from Qwen2.5-VL-32B outputs. For RL, 4,758 samples were selected from the training set, covering diverse cities, street scenes, and target types. The SFT stage is implemented using \textbf{LLaMA-Factory}~\cite{zheng2024llamafactory}, while the RL stage is built upon the \textbf{VLM-R1} framework~\cite{shen2025vlm}. Key hyperparameters for both stages are summarized in Table~\ref{tab:train-config}.



\begin{table}[t]
\small
\centering
\caption{Hyperparameters for SFT and RL stages.}
\label{tab:train-config}
\begin{tabularx}{\columnwidth}{lXXX}
\toprule
\textbf{Stage} & \textbf{Batch Size} & \textbf{LR} & \textbf{Epochs} \\
\midrule
SFT & 16 & 2e-5 & 2 \\
RL  & 1 & 1e-5 & 1 \\
\bottomrule
\end{tabularx}
\end{table}

\subsection{Experimental Results}
\subsubsection{Model Performance and Generalization Analysis}

Table~\ref{tab:all_result} summarizes the performance of various models across evaluation scenarios in the CityNav dataset.
Experimental results reveal that Qwen2.5-VL-7B achieves reasonable performance in UAV VLN tasks, while its larger variant, Qwen2.5-VL-32B, further improves and surpasses the strongest traditional baseline, MGP, across multiple metrics. These observations underscore that base VLMs already possess strong visual-language perception and multimodal fusion capabilities, even when used out-of-the-box without task-specific tuning.

On top of this foundation, FlightGPT further improves performance across the board. In the val-seen setting, it achieves the highest success rate \textbf{17.57\%}, the lowest navigation error \textbf{66.1}, and the most efficient path SPL \textbf{15.78}. In more challenging test-unseen setting, it shows remarkable generalization ability, improving the success rate by \textbf{9.22\%} and nearly doubling the SPL compared to Qwen2.5-VL-32B, the strongest baseline model.

It is worth noting that FlightGPT, built on the relatively lightweight Qwen2.5-VL-7B model, surpasses the larger-scale Qwen2.5-VL-32B after the application of a two-stage training pipeline. This result highlights that, rather than merely scaling up model size, incorporating appropriate modeling approaches (e.g., a CoT reasoning module) and adopting efficient training strategies (e.g., SFT+RL) are more crucial for enhancing model generalization and real-world performance.

\begin{table*}[t]
\centering
\tiny
\setlength{\tabcolsep}{2pt}
\caption{Comparison of Model Performance Across Evaluation Scenarios}
\label{tab:all_result}
\begin{tabular}{>{\centering\arraybackslash}p{3.2cm} | >{\centering\arraybackslash}p{0.77cm} >{\centering\arraybackslash}p{0.77cm} >{\centering\arraybackslash}p{0.77cm} >{\centering\arraybackslash}p{0.77cm} | >{\centering\arraybackslash}p{0.77cm} >{\centering\arraybackslash}p{0.77cm} >{\centering\arraybackslash}p{0.77cm} >{\centering\arraybackslash}p{0.77cm} | >{\centering\arraybackslash}p{0.77cm} >{\centering\arraybackslash}p{0.77cm} >{\centering\arraybackslash}p{0.77cm} >{\centering\arraybackslash}p{0.77cm}}
\toprule
\textbf{Method} & \multicolumn{4}{c|}{\textbf{Validation Seen}} & \multicolumn{4}{c|}{\textbf{Validation Unseen}} & \multicolumn{4}{c}{\textbf{Test Unseen}} \\
& NE$\downarrow$ & SR$\uparrow$ & OSR$\uparrow$ & SPL$\uparrow$ & NE$\downarrow$ & SR$\uparrow$ & OSR$\uparrow$ & SPL$\uparrow$ & NE$\downarrow$ & SR$\uparrow$ & OSR$\uparrow$ & SPL$\uparrow$ \\
\midrule
Random & 222.30 & 0.00 & 1.15 & 0.00 & 223.00 & 0.00 & 0.90 & 0.00 & 208.80 & 0.00 & 1.44 & 0.00 \\
Seq2Seq & 148.40 & 4.52 & 10.61 & 4.47 & 201.40 & 1.04 & 8.03 & 1.02 & 174.50 & 1.73 & 8.57 & 1.69 \\
CMA & 151.70 & 3.74 & 10.77 & 3.70 & 205.20 & 1.08 & 7.89 & 1.06 & 179.10 & 1.61 & 10.07 & 1.57 \\
MGP & 59.70 & 8.69 & \textbf{35.51} & 8.28 & 75.10 & 5.84 & 22.19 & 5.56 & 93.80 & 6.38 & 26.04 & 6.08 \\
Qwen2.5-VL-7B & 116.10 & 4.72 & 12.89 & 4.15 & 123.20 & 5.52 & 13.98 & 4.92 & 124.60 & 4.59 & 12.75 & 3.99 \\
Qwen2.5-VL-32B & 84.70 & 12.65 & 24.14 & 11.30 & 91.90 & 10.12 & 20.52 & 9.00 & 83.28 & 11.98 & 23.48 & 10.76 \\
LLaMA-3.2-11B-Vision & 198.90 & 1.16 & 5.16 & 1.06 & 215.10 & 0.50 & 4.35 & 0.46 & 191.10 & 1.26 & 4.59 & 1.15 \\
GPT-4o & 155.80 & 2.42 & 9.62 & 2.17 & 170.40 & 2.17 & 7.77 & 1.98 & 144.40 & 3.90 & 11.79 & 3.42 \\
SFT-only & 97.60 & 10.29 & 18.45 & 9.46 & 101.70 & 10.51 & 18.54 & 9.70 & 117.40 & 11.20 & 21.24 & 10.78 \\
RL-only & 74.90 & 13.27 & 27.13 & 12.59 & 71.40 & 12.87 & 25.82 & 12.27 & 76.50 & 19.80 & 32.26 & 18.91 \\
SFT+RL (FlightGPT) & \textbf{66.10} & \textbf{17.57} & 30.26 & \textbf{15.78} & \textbf{68.10} & \textbf{14.69} & \textbf{29.33} & \textbf{13.24} & \textbf{76.20} & \textbf{21.20} & \textbf{35.38} & \textbf{19.24} \\
\bottomrule
\end{tabular}
\end{table*}

\subsubsection{Ablation Study}
\begin{figure}
  \centering
  \includegraphics[width=0.97\columnwidth]{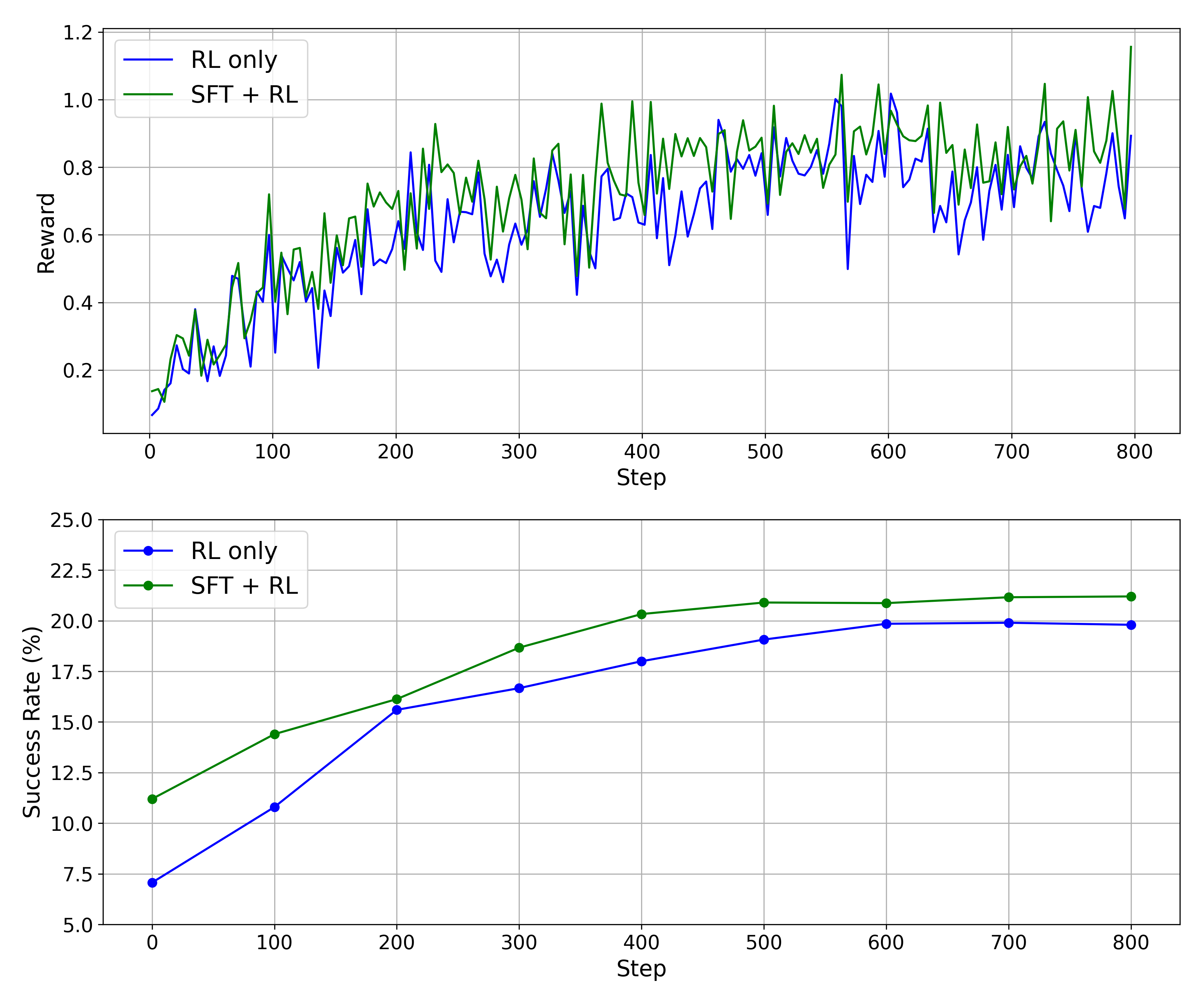} 
  \vspace{-8pt}
  \caption{Reward (train) and success rate (test) over training steps.}
  \label{fig:training_process}
\end{figure}

To systematically evaluate the contributions of SFT and RL in the FlightGPT framework, we conduct ablation experiments under the following three training configurations: (1) \textbf{SFT-only}: Trained with supervised fine-tuning only, without RL; (2) \textbf{RL-only}: Trained directly with reinforcement learning, without SFT initialization; (3) \textbf{SFT+RL}: Initialized with SFT and further optimized with RL.
\paragraph{Ablation Study Results Analysis.}
\begin{itemize}[leftmargin=*]
    \item \textbf{SFT-only}: This configuration achieves decent performance in the val-seen environment, benefiting from the reasoning mechanism and SFT on high-quality data. However, without RL for policy optimization and exploration, it shows limited generalization to OOD environments. On the test-unseen set, its performance is clearly inferior to models trained with RL.
    \item \textbf{RL-only}: This configuration eventually achieves reasonably good performance without any prior initialization. 
    However, as shown in Fig.~\ref{fig:training_process}, the model suffers from low success rates at the beginning of training due to the absence of a good starting policy. 
    Its convergence is slower than SFT+RL: while SFT+RL nearly converges at around 500 steps, the RL-only model only begins to stabilize after 600 steps, and its reward remains consistently lower throughout training. 
    In addition, its final performance remains slightly lower than that of SFT+RL.
    \item \textbf{SFT+RL}: The SFT stage provides a strong initialization of the policy, resulting in a more stable and faster convergence during training. 
    Subsequently, the RL stage further improves the model’s generalization and adaptability to OOD environments. 
    This configuration not only outperforms both SFT-only and RL-only baselines across all evaluation metrics, but also achieves a more stable and efficient training process, demonstrating the synergistic advantage of the two-stage training pipeline.
\end{itemize}

\subsubsection{Reasoning Quality Analysis}
To compare the reasoning quality between the RL-only and SFT+RL configurations, we randomly selected several cases from the dataset for qualitative analysis. 
The reasoning process generated by the RL-only model is generally disorganized: the \texttt{<think>} section tends to be short, lacks clear logical structure, and contains fragmented reasoning chains, making it difficult to follow. 
In contrast, SFT+RL produces significantly more coherent and well-structured reasoning, with complete chains covering landmark identification, spatial relation reasoning, and target location prediction. 
Several representative examples are provided in Appendix~\ref{app:reasoning-case-studies}.

To further quantify these observations, we designed three reasoning quality metrics and used GPT-4o to automatically score a random sample of 5,000 outputs (the detailed prompt is provided in Appendix~\ref{app:prompt_reasoning_quality}). The three reasoning quality metrics are measured by: (1) \textbf{Completeness}: Whether the reasoning covers all necessary steps without missing key details; (2) \textbf{Coherence}: Whether the reasoning is logically consistent and well connected throughout; (3) \textbf{Fluency}: Whether the language is fluent and grammatically correct.


To reduce evaluation variance, each sample was scored 3 times, and the average score was reported as the final result. The evaluation results, summarized in Table~\ref{tab:reasoning_quality}, show that SFT+RL outperforms the RL-only model across all three reasoning quality metrics, demonstrating the critical role of the SFT stage in improving reasoning quality. 
In particular, the SFT+RL configuration achieves a 0.44 improvement in completeness, indicating that structured reasoning training effectively guides the model to produce more comprehensive and systematic reasoning processes. 
Additionally, improvements of 0.26 and 0.08 are observed in coherence and fluency, respectively, further enhancing the clarity and readability of the reasoning outputs.




\begin{table}[t]
\small
\centering
\caption{Reasoning Quality Evaluation Results}
\label{tab:reasoning_quality}
\begin{tabular}{l>{\centering\arraybackslash}p{2.0cm} >{\centering\arraybackslash}p{1.5cm} >{\centering\arraybackslash}p{1.5cm}}
\toprule
\textbf{Strategy} & \textbf{Completeness} & \textbf{Coherence} & \textbf{Fluency} \\
\midrule
RL-only & 3.67 & 4.03 & 4.78 \\
SFT+RL & 4.11 & 4.29 & 4.86 \\
\bottomrule
\end{tabular}
\end{table}

\section{Conclusion}

In this paper, we propose FlightGPT, a system for UAV VLN, aiming to improve navigation performance in complex environments, enhance cross-environment generalization, and increase the interpretability of decision-making processes. 
We leverage the multimodal understanding capabilities of VLMs and construct a two-stage training pipeline that combines SFT with RL, where the RL stage is guided by a composite reward design.
In addition, we introduce a CoT reasoning mechanism to improve the transparency and controllability of the system.
We conduct comprehensive evaluations on the real-world, city-scale CityNav dataset. Experimental results show that FlightGPT achieves significant improvements over existing baseline models in in-distribution environments, and exhibits strong generalization capabilities in more challenging OOD scenarios.
We will release the code and data to facilitate further research.
\section{Limitations}
Despite the strong performance of FlightGPT in city-scale UAV VLN, several noteworthy limitations remain in terms of real-world applicability and system capabilities:

\textbf{Significant Gap Between Simulation and Reality.}
This study primarily relies on high-fidelity simulators such as CityNav for training and evaluation. While these platforms offer structured and diverse urban scenarios that facilitate learning of task semantics and spatial layouts, they fall short of capturing the complexity and uncertainty of real-world urban airspaces. Factors such as GPS drift, weather disturbances, dynamic obstacles, and unexpected events frequently arise in actual UAV operations and can significantly impact perception and decision-making. As a result, the system's performance, stability, and robustness in real-world settings remain unverified and call for further field testing and validation.

\textbf{Substantial Gap Compared to Human Navigation Abilities.}
Although FlightGPT demonstrates leading performance on the CityNav dataset and exhibits basic language understanding and path planning capabilities, its navigation intelligence still lags behind human operators. In particular, the model struggles with complex scenarios involving ambiguous expressions, implicit goals, or multi-turn instructions, often lacking commonsense reasoning and strategic flexibility. This exposes limitations in multi-modal semantic integration, spatial reasoning, and decision consistency, making it difficult for the system to handle dynamic and high-complexity navigation tasks.

\textbf{Lack of Systematic Evaluation of Deployment Feasibility.}
The current research primarily emphasizes performance, with insufficient attention paid to the practical requirements for real-world deployment. Key factors such as inference latency, memory usage, and computational resource demands directly influence the system's ability to operate in real time on resource-constrained edge devices, yet these metrics have not been systematically quantified. Furthermore, issues such as communication reliability and failure recovery mechanisms—critical for engineering-level implementation—remain underexplored, limiting the transition of FlightGPT from a research prototype to a deployable solution.

\section{Broader Impact and Ethics}
\textbf{Dual-use risk.}
UAV-based navigation systems, while beneficial for disaster relief or infrastructure inspection, may also be misused for surveillance, tracking, or other purposes that infringe on privacy or civil liberties. To mitigate such risks, real-world deployment should be accompanied by appropriate regulatory oversight, strict usage boundaries, and human-in-the-loop supervision mechanisms.

\textbf{Risk of unsafe deployment.}
Although the system shows strong performance in simulated city-scale environments, deploying it in real-world scenarios poses safety risks due to unmodeled factors such as GPS drift, occlusions, dynamic obstacles, or weather conditions. Without rigorous field testing and fail-safe mechanisms, these issues may lead to unintended navigation failures or even physical harm to people or property.

\newpage

\bibliography{custom}

\newpage
\appendix

\section{Prompt Template for FlightGPT}
\label{app:prompt}

\begin{prompt}{Prompt}
\textbf{\small System Message:}

You are an intelligent autonomous aerial vehicle (UAV) capable of real-world navigation and visual target localization.
\\
\\    
\textbf{\small Mission Objective:}

Your mission is to locate a specific target described in natural language instructions.
\\
\\
\textbf{\small Details of the Target:}

\verb|{target description}|
\\
\\
\textbf{\small Environmental Perception:}

    - The UAV's current position is indicated by the starting point of an arrow in the image, with its heading angle represented by the arrow's direction. \\
    - The yellow box outlines the UAV's current camera field of view on the map, centered at pixel coordinates: \verb|cur_pose = {UAV current position}|. \\
    - Landmark regions are highlighted with red masks.
\\
\\
\textbf{\small Operational Guidance:}

    - The target is usually located near a red-masked landmark. \\
    - Use both the target description and the visual input to identify the most relevant red-masked landmark region. \\
    - Infer the relative position of the target with respect to that landmark.
\\
\\
\textbf{\small Output Format Specification:}

    - Present your reasoning process within \texttt{<think>} and \texttt{</think>} tags.\\ 
    - Provide your final answer within \texttt{<answer>} and \texttt{</answer>} tags in the following format: \verb|{"target_location": [x, y]}| \\
    Your reasoning may include: \\
          $\quad$- A semantic interpretation of the target description. \\
          $\quad$- Identification of the correct landmark region. \\
          $\quad$- The bounding box of that region in the following format: \\
          \verb|{"landmark_bbox": [x1, y1, x2, y2]}|
\end{prompt}

\section{Baseline Model Descriptions}
\label{app:baselines}
We briefly introduce the baseline models evaluated on the CityNav.
\begin{itemize}[leftmargin=*, itemsep=0pt]
    \item \textbf{Random}: A simple random action policy serving as a lower bound reference.
    
    \item \textbf{Seq2Seq} \cite{2017Vision}: A classic end-to-end model that encodes the instruction and visual input into a latent representation to directly generate action sequences.
    
    \item \textbf{CMA} \cite{0Generating}: A cross-modal attention model that integrates visual and linguistic inputs via cross-attention mechanisms for action selection.
    
    \item \textbf{MGP} \cite{lee2024citynavlanguagegoalaerialnavigation}: A map-based method that predicts the target location based on semantic maps and natural language instructions.
    
    \item \textbf{GPT-4o} \citep{openai2024gpt4ocard}: A powerful multimodal model developed by OpenAI, capable of processing both text and visual inputs with reasoning abilities.
    
    \item \textbf{Qwen2.5-VL (7B / 32B)} \cite{bai2025qwen25vltechnicalreport}: An open-source family of vision-language models with varying parameter scales, used to assess how model size influences performance.
    
    \item \textbf{LLaMA-3.2-11B-Vision} \cite{grattafiori2024llama}: Meta AI's latest open-source multimodal model, supporting both visual and language perception.
    
\end{itemize}


\section{Reasoning Examples}
\label{app:reasoning-case-studies}

\begin{table*}[t]
\scriptsize
\centering
\caption{Reasoning Examples}
\label{tab:appendix_case_studies}
\begin{tabular}{c p{6cm} p{8cm}}
\toprule
\textbf{ID} & \textbf{RL-only Output (<think> Section)} & \textbf{SFT+RL Output (<think> Section)} \\
\midrule
1 & 
The landmark region is the area where the red mask is present. The bounding box for the red-masked region is:

\texttt{\{"landmark\_bbox": [3428, 23, 3801, 547]\}}

The probable target location is on the sidewalk next to the building within this red-masked area. 
& 
The target is a black car on the sidewalk next to a building on Chesterwood Gardens. The red masks in the image indicate the streets, specifically Chesterwood Gardens, and may help us identify the correct landmark region for locating the target. 

\textbf{1. Semantic Interpretation of the Instruction:} 

-- The car is described as a black car and is on the sidewalk near a building. 

-- The location reference ``Chesterwood Gardens'' should point the drone towards the correct residential area.

\textbf{2. Identifying the Correct Landmark Region:}

-- The aerial view shows Chesterwood Gardens as a rectangular area marked by the red mask. Given it is residential with houses and likely the target is near the edge of one of these buildings on the sidewalk, I would focus on the northern edge of Chesterwood Gardens.

\textbf{3. Relative Position of the Target:}

-- The target is described as on the sidewalk next to a building. The sidewalk along the northern edge of Chesterwood Gardens is a likely location.

\textbf{4. Bounding Box for the Landmark Region:}
-- Based on the aerial view, the northern edge of Chesterwood Gardens is highlighted in red. The area around the edge of the buildings and the sidewalk here aligns with the target description.

\texttt{\{"landmark\_bbox": [2693, 16, 4026, 1361]\}}
\\
\midrule
2 & 
The target, a black car parked between a silver car and a black car, is located in the parking lot behind the "One Stop" store. The red-masked landmark represents the One Stop store, which is a good reference point. The target is likely to be in the parking lot area adjacent to this landmark, as indicated by the "One Stop" text.

The parking lot is clearly demarcated by the red masks. Given the position of the target relative to the One Stop landmark, it should be in the area where the parking rows are arranged around the store. Since the target is described as being between two cars, we can look for that specific arrangement of cars in the parking lot area behind One Stop.

Since the target is in the parking lot area behind One Stop, the bounding box would cover the parking lot region near the red-masked "One Stop" area.

\texttt{\{"landmark\_bbox": [1745, 0, 3882, 1467]\}}

The target, a black car, is likely to be found within the bounding box of the parking lot area behind "One Stop".
& 
The target description states that it is a black car parked between a silver car and a black car in a lot behind "One Stop." Given the current position indicated by the arrow, we need to identify a red-masked landmark near this area as the target might be near a red-masked street landmark for reference.

\textbf{1. Identify Possible Red-Masked Landmarks:} 

-- The image highlights red structures with some marked as "One Stop," which appears larger and more prominent.

-- The area behind "One Stop" is clearly outlined by the red mask.

\textbf{2. Locate the Target Relatively:}

-- The target is described as being behind "One Stop." 

-- Given the current field of view and the position indicated by the arrow, the target should be within the vicinity of the "One Stop" landmark.

\textbf{3. Bounding Box for Landmark:}

-- The "One Stop" landmark is clearly marked by the red structure.

-- The bounding box for the landmark can be identified roughly around the "One Stop" structure.

\texttt{\{"landmark\_bbox": [1594, 0, 3864, 1502]\}}
\\
\midrule
3 
& 
The target description indicates that the target is a parking lot located behind the building at the corner of Chesterton Road and Ferry Path. The red mask outlines the street landmark regions. The target should be in the parking lot behind the building at the intersection of Chesterton Road and Ferry Path.

Let's identify the landmark region:

-- The red mask around Chesterton Road and Ferry Path clearly highlights the region where the target is likely to be located.

-- The target parking lot is behind the building at the corner of Chesterton Road and Ferry Path.

The bounding box for the landmark region is:

\texttt{\{"landmark\_bbox": [403, 249, 1060, 959]\}}

Given the green arrow and the current orientation of the UAV, the target is likely to be behind the building at the corner of Chesterton Road and Ferry Path within the red mask area.
& 
The task is to find a parking lot, which is described as "behind the building at the corner of Chesterton Road and Ferry Path" and "has no cars in this area." The goal is to locate this parking lot, relative to the red-masked landmark regions marked as "Chesterton Road" and "Ferry Path".

\textbf{1. Identify the relevant landmark red masks:}

-- There are red masks over Chesterton Road and Ferry Path in the image. These identify the streets and regions of interest.

\textbf{2. Determine the direction of movement:}

-- The UAV's current orientation is indicated by the arrow at the bottom-left. It points towards the top-left of the image.

\textbf{3. Locate the intersection:}

-- The intersection of Chesterton Road and Ferry Path is where the red masks overlap at the bottom of the image.

\textbf{4. Identify the target area:}

-- The target parking lot is described as being behind the buildings at the corner. Given the red masks, it's reasonable to conclude that any area behind these masks, specifically in the region near the intersection, is the location to look for.

\textbf{5. Bounding box of the area:}

-- Since we need to look behind the intersection at the parking lot, and considering the red masks as a guide, I'll assume an area directly behind the intersection is our target. This means the parking lot would lie inside the yellow bounding box (highlighted on the image).

The bounding box region of interest behind the red-masked Chesterton Road, which should contain the parking lot, would be:

\texttt{\{"landmark\_bbox": [370, 273, 1076, 972]\}}
\\
\bottomrule
\end{tabular}
\end{table*}

\section{Prompt for Reasoning Quality Evaluation}
\label{app:prompt_reasoning_quality}

\begin{prompt}{Prompt}
\textbf{\small System Message:}

You are a language model output quality evaluator. Please evaluate the following text based on the three criteria below.
\\
\\    
\textbf{\small Evaluation Criteria Definitions:}

    1. Completeness  
    
    - Does the text fully present all the necessary reasoning steps or information needed to answer the question or solve the problem?  
    
    - Are there any missing steps, skipped logic, or gaps in reasoning? \\
    2. Coherence  
    
    - Is the reasoning internally consistent and logically connected? Are there any contradictions or abrupt jumps?\\
    3. Fluency  
    
    - Is the language natural and smooth? Are grammar, sentence structure, and word choice appropriate?
\\
\\
\textbf{\small Scoring Guideline:}

    - 1: Very poor  \\
    - 2: Poor  \\
    - 3: Fair  \\
    - 4: Good  \\
    - 5: Excellent  \\
    Based on the above definitions and scoring scale, please evaluate the following text. Output one integer score (1–5) per criterion. Strictly follow the format below so it can be parsed by code.
\\
\\
\textbf{\small Language Model Output:}

\verb|{language model output}|
\\
\\
\textbf{\small Output Format:}

    - Completeness: x \\
    - Coherence: x\\
    - Fluency: x 
\end{prompt}

\end{document}